\documentclass[10pt,twocolumn,letterpaper]{article}

\usepackage[pagenumbers]{cvpr} 

\usepackage{graphicx}
\usepackage{amsmath}
\usepackage{amssymb}
\usepackage{booktabs}
\usepackage{multirow}

\usepackage[dvipsnames]{xcolor}

\usepackage[pagebackref,breaklinks,colorlinks]{hyperref}

\usepackage[capitalize]{cleveref}
\crefname{section}{Sec.}{Secs.}
\Crefname{section}{Section}{Sections}
\Crefname{table}{Table}{Tables}
\crefname{table}{Tab.}{Tabs.}

\def\cvprPaperID{6787} 
\def\confName{CVPR}
\def\confYear{2023}

\begin{document}

\title{Making the Most of What You Have:\\ Adapting Pre-trained Visual Language Models in the Low-data Regime}

\author{Chuhan Zhang$^1$ \quad Antoine Miech$^2$ \quad Jiajun Shen$^2$ \quad Jean-Baptiste Alayrac$^2$ \quad Pauline Luc$^2$\\
VGG, University of Oxford$^1$ \quad DeepMind, London$^2$\\
{\tt\small czhang@robots.ox.ac.uk\quad \{miech,jiajuns,jalayrac,paulineluc\}@google.com}
}

\maketitle

\begin{abstract}
Large-scale visual language models are widely used as pre-trained models and then adapted for various downstream tasks. 
While humans are known to efficiently learn new tasks from a few examples, deep learning models struggle with adaptation from few examples. 
In this work, we look into task adaptation in the low-data regime, and provide a thorough study of the existing adaptation methods for generative Visual Language Models. 
And we show important benefits of self-labelling, i.e.\ using the model's own predictions to self-improve when having access to a larger number of unlabelled images of the same distribution.
Our study demonstrates significant gains using our proposed task adaptation pipeline across a wide range of visual language tasks such as visual classification (ImageNet), visual captioning (COCO), detailed visual captioning (Localised Narratives) and visual question answering (VQAv2).
\end{abstract}


\section{Introduction}
\begin{figure}[t]
\centering
\includegraphics[width=1.02\linewidth]{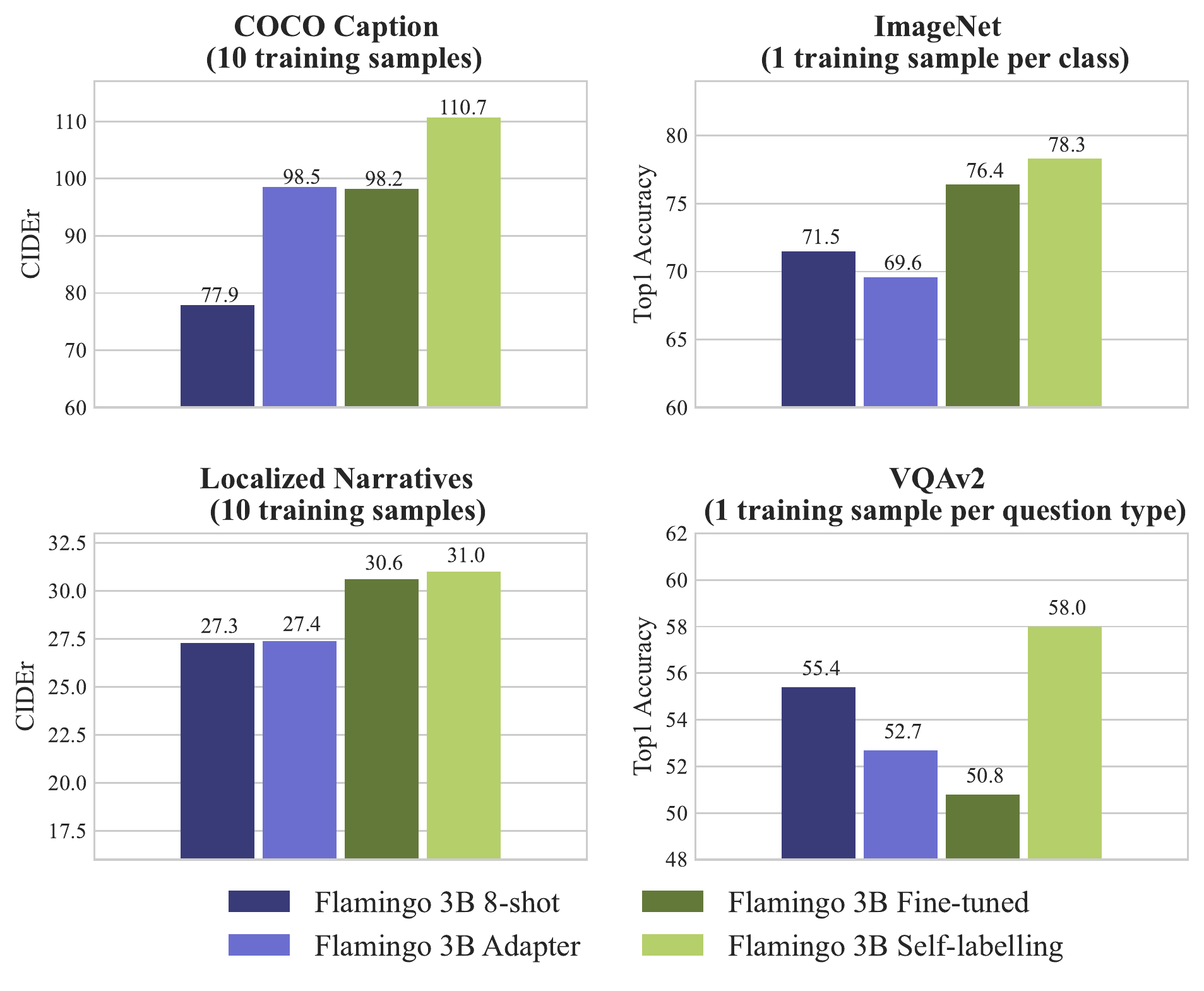}
\caption{\textbf{Adapting pre-trained visual language model to new tasks in the low-data regime.} We compare different methods for adapting a VLM to image captioning, classification and VQA.  We show that simple {\color{YellowGreen}\textbf{self-labelling}} improves the performance on target tasks consistently on four datasets: COCO, Localized Narratives, ImageNet and VQAv2. More detailed results are presented in \cref{tab:results}.}
\end{figure}

Humans are able to perform novel tasks given only a handful of examples. 
For example, having only seen a few instances of a new class of objects, humans are generally able to recognize other instances of this class, despite changes in appearance or pose.

Such a property is desirable for machine learning systems since collecting large number of manual annotations is tedious, expensive and unrealistic to do in practice for all tasks that one may want to automate. 
As such, studying better ways to adapt existing models to novel tasks in the extreme low data regime is of utmost importance.

In recent work, Visual Language Models (VLMs) have shown great success in a variety of visual tasks\cite{alayrac2022flamingo,li2022blip,clip2021,yuan2021florence}. VLMs take a sequence of images and text as input and can be used to generate or score a text sequence. They are pre-trained on large-scale vision and language data, therefore can be evaluated on various downstream tasks which are similar to the pre-training tasks. Flamingo~\cite{alayrac2022flamingo} shows that the pre-trained model can be adapted to new tasks with in-context learning, where the few-shot samples and the query are provided to the model in an interleaved format as a prompt of images, the model is used to generate scores or new-form answers following the examples in the prompt. 
In this work, we propose to go beyond in-context learning as a task adaptation method, and ask how best to adapt to a new task from as few as 10 annotated images.

Addressing this question is challenging for multiple reasons.
First, large models are prone to over-fitting in such extreme low data regime, since the number of parameters is much bigger ($\approx$billions) than the number of annotated data points ($\approx$tens).
Second, a single adaptation method may exhibit varying performance across different tasks, depending on the difference between the output formats.
To tackle the aforementioned challenges, we conduct a comprehensive study on the impact of diverse adaptation methods on multiple tasks, as well as explore ways to overcome the issue of limited annotated data across all the tasks.

We start from Flamingo~\cite{alayrac2022flamingo}, a powerful pre-trained Visual Language Model which has good generalization capability in the few-shot data regime. We propose a pipeline to adapt this model to a novel task given only a small amount of annotations.
This pipeline consists of three stages: \textbf{(1)} train a pseudo-labeller, \textbf{(2)} use it to obtain pseudo-labels and then \textbf{(3)} re-train the original model using the pseudo-labels and the small amount of annotated data.

We first explore multiple strategies (stage 1) to perform task adaptation from only a few annotated data points. This includes methods such as in-context learning, where the model is prompted with the examples of the task, adapter-based method~\cite{houlsby2019parameter}, and also regular model fine-tuning.
This study is conducted on diverse image tasks: captioning (with COCOCap~\cite{cococap2015}), object classification (with ImageNet~\cite{imagenet}), visual question answering (with VQAv2~\cite{balanced_vqa_v2}) and detailed image captioning (with LocNar).
Given a strong adapted model, we then explore how to best use it as a pseudo-labeller to further improve the model predictions when given a large set of unlabelled images from the same input distribution as the task of interest, in a semi-supervised regime.
We notably demonstrate significant gains by using the adapted model's own prediction to obtain pseudo-labels (stage 2) and carefully incorporating them into semi-supervised training (stage 3).

\noindent
\textbf{Contributions.} Our contributions are three fold:
\textbf{(i)} We give a thorough analysis of different adaptation techniques for state-of-the-art visual language models in the (very) low data regime (stage 1 of the pipeline),
\textbf{(ii)} we then show how we can use the best adapted model to self-label a large amount of unlabelled images
of the same distribution and how to use these pseudo-labels to improve the initial model performance (stage 2 and 3),
\textbf{(iii)} We compare the self-labelling results against different sizes of self-labelled data;
\textbf{(iv)} Finally, we showcase the effectiveness of the proposed approach through a experimental study on 4 different tasks (ImageNet, COCOCap, Localized Narrative and VQAv2), demonstrating substantial 
performance improvements from as few as 10 annotated examples.

\noindent
\textbf{Summary of findings.}
Our results can be summarized in three main findings, illustrated in the rest of the paper and summarized next.
First, the effectiveness of different adaptation methods is task-dependent. With very little data (as few as 10 training images), using fine-tuning is overall the most competitive method on captioning and classification, while in-context learning performs better on VQA.

Second, we observe that training with self-labelled data from the best adapted model brings further gains across all our tasks.
Finally, we also note that when using a large-scale pre-trained model, filtering and heavy augmentation is not critical for performance. Simple semi-supervised training is sufficient to improve the performance on target tasks.

\begin{figure*}[t]
\centering
\includegraphics[width=0.93\linewidth]{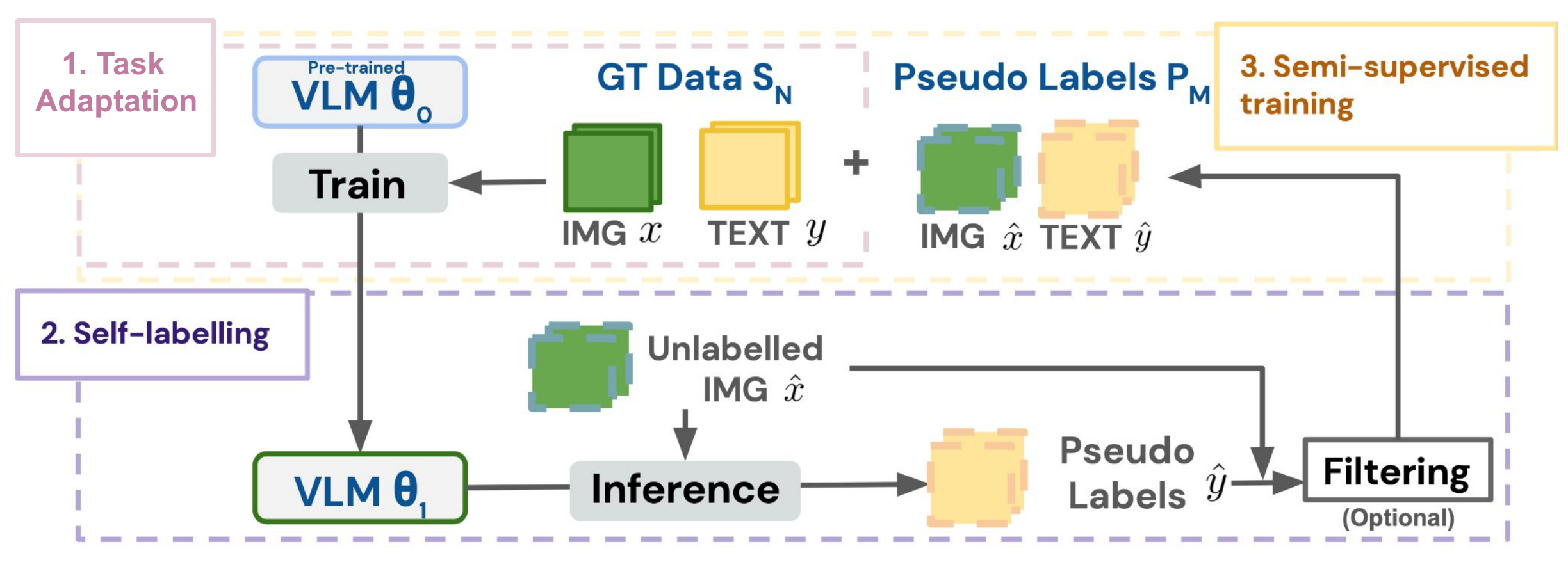}
\caption{\label{fig:pipeline}\textbf{Pipeline for model adaptation on few annotated data.} This figure illustrates the three stages of our pipeline. \textbf{1. Adaptation:} Given a large-scale pre-trained visual language model~$\theta_{0}$, we first obtain a model $\theta_{1}$ from a small number of annotated data $S_{N}$. 
\textbf{2. Self-labelling:} The trained model $\theta_{1}$ is used as a pseudo-labeller to annotate a large number of unlabelled images from the same data source, followed by a filtering stage to reduce label noise and obtain our final pseudo-label set $P_{M}$.
\textbf{3. Semi-supervised training:} We fine-tine the model $\theta_{0}$ on a mixture of data composed of $S_{N} + P_{M}$ by carefully weighting the relative importance of both sets.}
\end{figure*}

\section{Related Work}

In this section, we review related work and how they differ from ours.

\subsection{Visual Language Models}

We focus our study on vision and text.
The large number of vision and text models can be categorized in three main families.
First, we have the BERT~\cite{bert} inspired family of vision-text model~\cite{lu2019vilbert,su2019vl,chen2020uniter,hendricks2021decoupling,wang2021vlmo,li2020oscar,tan2019lxmert,zhu2020actbert,wang2021ufo,li2020hero,gan2020large,fu2021violet,zellers2021merlot,zellers2022merlot,singh2021flava}.
These works aim at training a joint vision-text representation using  masked language modelling and / or a masked image region modelling loss.
The downside of these approaches is that they often need to be fine-tuned on new tasks using a large-number of annotated examples.
Second, we have the family of contrastive vision-text models~\cite{alayrac2020self,clip,align,zhai2021lit,pham2021combined,miech2020end,bain2021frozen,yuan2021florence,li2021align,yao2021filip,jain2021mural}.
As opposed to the BERT-like vision-text models, contrastive vision-text models are known to be quickly adapted to tasks such as classification or retrieval in zero-shot, using adequate prompting~\cite{clip}.
However, it is not straightforward to adapt contrastive models for generative tasks such as visual captioning or visual question answering.
Instead, our work relies on the family of generative, visual language models~\cite{alayrac2022flamingo,chen2021visualgpt,desai2021virtex,eichenberg2021magma,vinyals2015show,donahue2015long,luo2020univl,hu2021scaling,wang2021simvlm,cho2021unifying,wang2022unifying,zhu2021uni,li2022blip}.
The vast majority of these models are trained with an auto-regressive language modeling loss.
This allows them to generate text and can thus be adapted to a wider ranger of tasks ranging from classification, retrieval, question answering to captioning.
In particular, we use the recent state-of-the-art Flamingo 3B~\cite{alayrac2022flamingo} visual language model, known for its ability to be quickly adapted to a wide variety of tasks.
One additional benefit of generative models is that it can be used as a generator of pseudo-labels in our self-labelling stage.

\subsection{Adapting pre-trained models to new tasks}

Our work focuses on the data-efficient adaptation of models to different vision-text tasks.
The common approach for adapting vision-text models is to fine-tune all their weights using gradient-based techniques~\cite{robustfinetune2022,clip4clip2022,ilharco2022patching}.
However, this is commonly considered as not being data-efficient as it often requires tuning an order-of-magnitude more parameters than available training examples.
To address this, several work instead propose to only fine-tune a small subset of the model weights~\cite{chen2021visualgpt,eichenberg2021magma,tsimpoukelli2021multimodal}.
Other work instead introduce a relative low number of trainable weights, tuned from scratch on new tasks.
These include prefix / prompt-tuning~\cite{zhou2021learning,li2021prefix,lester2021power,zhu2021uni},
bottlenecked MLP-Adapters~\cite{adapter2019,eichenberg2021magma,rebuffi2018efficient,multitaskadapter2022} or convolutional adapter layers~\cite{rosenfeld2018incremental}.
More recently, inspired by the successes of few-shot in-context learning in language~\cite{gpt3}, in-context learning in vision and text has been proven to be effective at quickly adapting with as few as four training examples~\cite{tsimpoukelli2021multimodal,alayrac2022flamingo}.
Our work compares many of these data-efficient adaptation techniques on a diverse set of tasks.

\subsection{Self-training and pseudo-labelling}
Self-training has been widely studied as an efficient way to improve the performance of models with noisy pseudo-labels produced by itself. For example,~\cite{lee2013pseudo,yalniz2019billion,noisystudent2020,sohn2020fixmatch,pham2021meta,arazo2020pseudo} have shown that by utilising self-labelled images, better image classification accuracy can be achieved. Similar conclusions are also reported in many studies from different domains, showing self-training can be successfully applied to improve state-of-the-art performance on many tasks, including~\cite{xie2020unsupervised,He2020Revisiting,kahn2020self,Park2020}. 
These works typically train a teacher model on labelled data to generate pseudo-labels, and then train the student model on the combination of labelled and pseudo-labelled data. Heavy augmentations are often applied on pseudo-labels, and it is shown to be helpful for improving robustness and enforcing model consistency~\cite{noisystudent2020,sohn2020fixmatch}. 

Recent work \cite{chen2020big} looks at self-training with a self-supervised vision model, and shows that big self-supervised models are strong semi-supervised learners which can learn efficiently from unlabelled images. The same explorations have been done on language model as well, \cite{selfimprove2022} shows that large language model can self-improve on open-ended QA tasks. With the recent development of generative visual language models~\cite{alayrac2022flamingo,chen2021visualgpt,desai2021virtex,eichenberg2021magma,vinyals2015show}, self-labelling has also been used to bootstrap the performance on image captioning~\cite{blip2022}. Previous works show the advantage of self-training, but focus mostly on self-labelling on a specific task, and requires the model to be trained on a large number of labelled data from the corresponding task. In this work, we want to utilize both the large-scale pre-training and the versatility of a large VLM on different tasks, and investigate how much self-training can help on multiple tasks with only a few labels from each.

\section{Learning in the low data regime}\label{sec:ft_method}

Given a small number of annotations of a novel task and a visual language model pre-trained on large-scale weakly annotated web data, our objective is to best adapt the model to the target task.
In this work, visual language model refers to a visually conditioned language model which is able to predict the likelihood of text given a sequence of interleaved text and images and from which language can be generated.
We consider here a diverse set of tasks, from open-ended tasks like captioning or VQA, to closed-ended tasks like classification. 
In all the tasks, we assume that the model only has access to a limited number of annotations for training and validation. 
More specifically, we make the assumption that there are only $N$ training images labeled with text descriptions for the captioning task ($N\in \{10, 100, 1000, 10000\}$), $N$ training images per class for the classification task ($N\in \{1, 5, 10, 15\}$), and $N$ training images per question type for text-conditioned tasks like VQA ($N\in \{1, 50, 100, 150\}$). These training samples are selected randomly from the original training set. We also reserve $M$ images from the original training set for validation ($M=200$) and conduct testing on the full test set.
We give an overview of our proposed pipeline for task adaptation in Section~\ref{sec:pipeline} before describing in more details the stages of the pipeline in Section~\ref{sec:adapting} and Section~\ref{sec:selflabelling}.

\subsection{Pipeline}
\label{sec:pipeline}

We here provide an overview of our overall pipeline for task adaptation. This process is also illustrated in Figure~\ref{fig:pipeline}.

\paragraph{Stage 1: Training a pseudo-labeller.} 
We first fine-tune or few-shot prompt a generative visual language model $\theta_{0}$ with a fixed number of ground-truth image and text pairs $ S_{N} = \{ (x_1, y_1), (x_2,y_2), ..., (x_N,y_N)\} $. 
In the case where some parts of the model are fine-tuned or new parameters are introduced, we use the negative log likelihood loss~\cref{nll_loss} to obtain a pseudo-labeller with updated parameters $\theta_{1}$:
\begin{equation}
    \mathcal{L}_{\text{NLL}} (\theta, S_{N}) = - \sum_{i=1, x,y \in S_{N}}^{N} {\log(P_{\theta}{(y_i \vert x_{i})}}).
\label{nll_loss}
\end{equation}

\paragraph{Stage 2: Self-labelling.} 
We use the previously trained pseudo-labeller with parameters $\theta_{1}$ to annotate the unlabelled images $\{ \hat{x}_1, \hat{x}_2, ..., \hat{x}_M \} $ from the same domain and obtain pseudo-labels $\{ \hat{y}_1, \hat{y}_2, ..., \hat{y}_M \} $, to construct a new pseudo-labelled dataset $P_{M} = \{ (\hat{x}_1, \hat{y}_1), (\hat{x}_2, \hat{y}_2) ..., (\hat{x}_M, \hat{y}_M) \}$.


\paragraph{Stage 3: Semi-supervised training.}
We re-train model $\theta_{0}$  with both the GT labels $y$ and pseudo-labels $\hat{y}$ and minimize the combined loss $\mathcal{L}$: 

\begin{equation}
    \mathcal{L}(S_{N},P_{M}) = \alpha  \mathcal{L}_{\text{NLL}} (\theta, S_{N}) + (1- \alpha ) \mathcal{L}_{\text{NLL}} (\theta, P_{M}),
\label{eq:cotrain_loss}
\end{equation} 
where $\alpha\in\mathbb{R}$ is a hyper-parameter used to balance the small amount of annotated data against the large amount of pseudo-labelled data during the semi-supervised learning stage.

\subsection{Training the most data efficient pseudo-labeller}\label{sec:adapting}

We review below the three main methods we consider to adapt a model given a few annotations: fine-tuning, adapter based methods and in-context learning.

\paragraph{Fine-tuning.} 
This is the most common way of adapting a model to a new task.
It effectively adapts the model to the new data distribution by changing the model weights.
The optimization of large number of parameters allows it to handle large domain gap between the source task and target task effectively, leading to good adaptation performance.
However, it is prone to over-fitting when the number of training data is small compare to the size of the model.
In our experiments where we use Flamingo-3B, we only fine-tune the layers that are not frozen during Flamingo main training \ie the Perceiver Resampler and the interleaved cross-attention dense layers.

\paragraph{Adapter.}
Different from fine-tuning, the adapter approach consists in adding a few additional modules where parameters are initialized from scratch and only train these ``adapters'' on the new task by keeping all the previously trained parameters unchanged.
The advantage is that there is a smaller number of parameters to train.
In details, we follow the approach in~\cite{adapter2019} and add MLP adapter layers to all the self-attention, cross-attention and feed-forward layers in the Flamingo model.
These adapter layers are then trained from scratch along with the pre-existing Layer Normalization parameters~\cite{ba2016layer} while all other parameters inside the model are kept frozen.

\paragraph{In-context learning.}
In-context learning has been proposed as another approach to task adaptation for visual language models~\cite{alayrac2022flamingo}. 
It adapts the model rapidly to new tasks by few-shot prompting, without the need to train the model again. 
In detail, the few-shot prompt is obtained by concatenating a set of support images $\{ x_{1},x_2, ..., x_{l-1}\} $ and textual annotations $\{ y_{1},y_2, ..., y_{l-1}\} $ from the new task and a single visual query from the new task.
During inference, the model then produces a prediction $\hat{y}_l$ on the query image $x_{l}$ conditioned on the examples in the support set.
To obtain the best performance from few-shot prompt, we follow~\cite{alayrac2022flamingo} and use retrieval-based in-context example selection (RICES)~\cite{rices2022} on classification tasks. 
Although in-context learning does not involve updates to the model parameters, RICES allows the model to benefit from more training samples, as the likelihood of sampling similar images is increased when using a bigger support set.

\subsection{Obtaining and selecting the pseudo-labels}
\label{sec:selflabelling}

\subsubsection{Pseudo-Labelling on different tasks}
\paragraph{Captioning and Classification.} 
In open-ended tasks like captioning, there are usually no constraints on the output space. 
Given an image and label as the input, we simply take what the model predicts as pseudo-labels. 
Specifically, we use beam search as decoding method for generation. 
In closed-ended tasks such as classification, predictions are constrained to be within a pre-defined set of classes, where we compute the likelihood of all the possible classes and choose the one which maximizes the likelihood.

\begin{figure}[b]
\includegraphics[width=\linewidth]{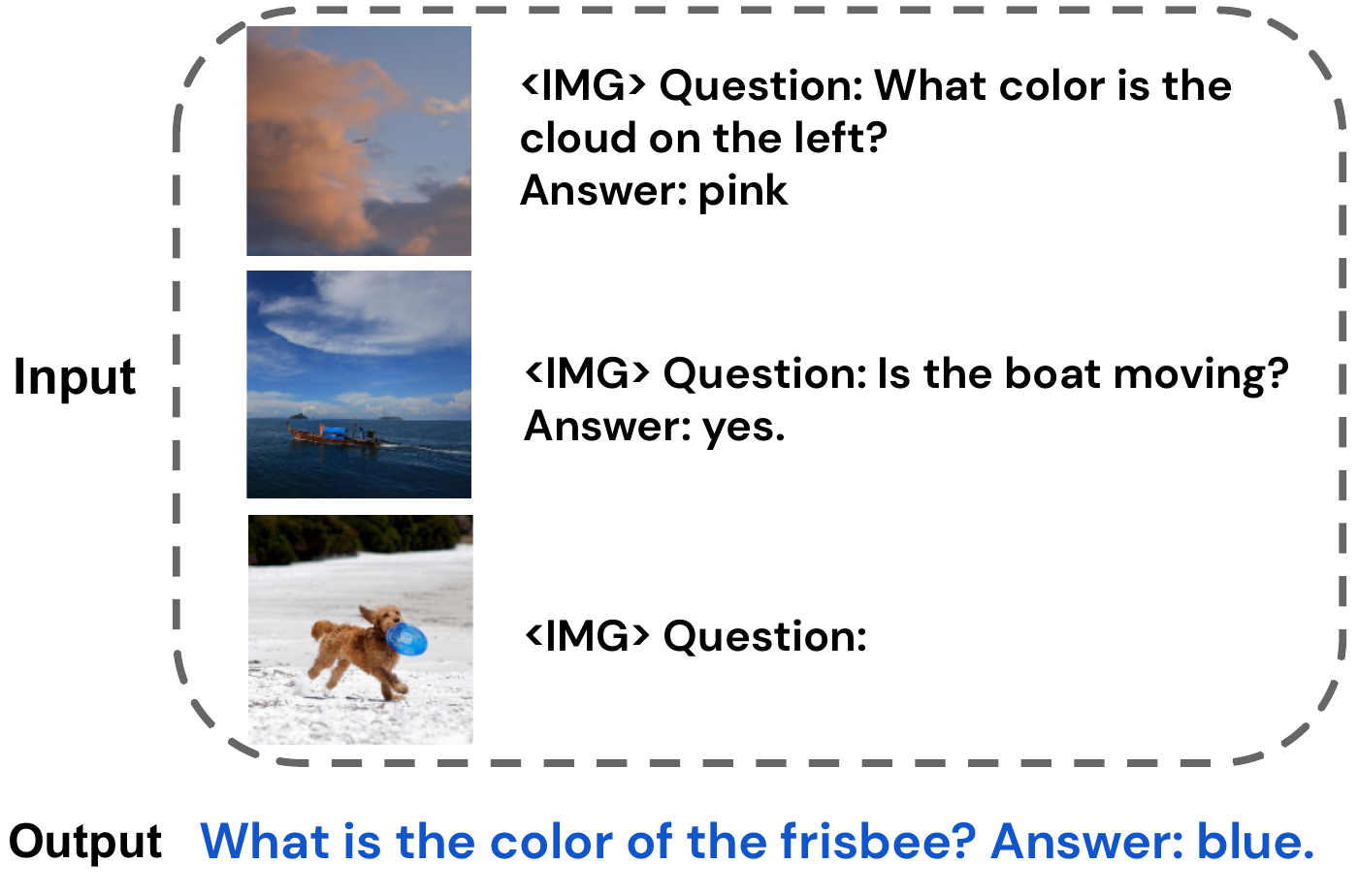}
\caption{\textbf{Generating questions and answers for VQA through few-shot prompting.} We concatenate the images and their corresponding QA pairs from the manually-labelled training data as the few-shot support set, and prompt the model to output question and answer given a query image.}
\label{fig:fewshot-qa}
\end{figure}

\paragraph{VQA.}
Different from captioning and classification which only take images as input, VQA takes a question in addition to the image. Since only unlabelled images are available in the self-labelling process, we need a way to generate the question and answer given an image. 
For instance, one way to do this is to leverage the in-context learning ability from Flamingo~\cite{alayrac2022flamingo} to enable the generation of both question and answer. As shown in \cref{fig:fewshot-qa}, by concatenating a few support images and their annotated QA pairs  with the query image and a text prompt `Question:' to form the input,
the model correctly outputs a pair of question and answer in the same format that are relevant to the query image.

\subsubsection{Filtering techniques} \label{sec:filtering}
The pseudo-labels produced can be wrong, thus filtering is used to improve the quality of pseudo-labels and obtain better results in semi-supervised learning.
For this reason, we explore filtering approaches commonly used for that purpose in the literature as detailed next.

\paragraph{Contrastive filtering.}
Vision language contrastive models~\cite{clip2021,blip2022} are often used to measure the similarity between images and texts. In this work, we carry out contrastive filtering in two steps: \textbf{1)} For each image, selecting the predicted text with the highest similarity score from top $K$ beam search results, and \textbf{2)} among all the images annotated by the model, selecting $N$ samples with the highest similarity scores.

\paragraph{Confidence filtering.}
The estimated probability output by the model can also be used as a filtering approach~\cite{blip2022,noisystudent2020}. 
We also explore filtering based on the negative likelihood of the predicted strings. 
We rank the predictions from beam search by likelihood, and choose the top $N$ samples with the highest likelihood.

\section{Experiments}

In this section, we start by introducing datasets in~\cref{sec:dataset}, followed by implementation details in~\cref{sec:imple}.
We then ablate the choice of hyper-parameters, filtering and training with pseudo-labels in \cref{sec:ablation}. 
Informed by this study, we then present our final results obtained on the test sets of the different tasks we consider in \cref{sec:results}.
\begin{figure*}[t]
\minipage{0.25\textwidth}
  \includegraphics[width=\linewidth]{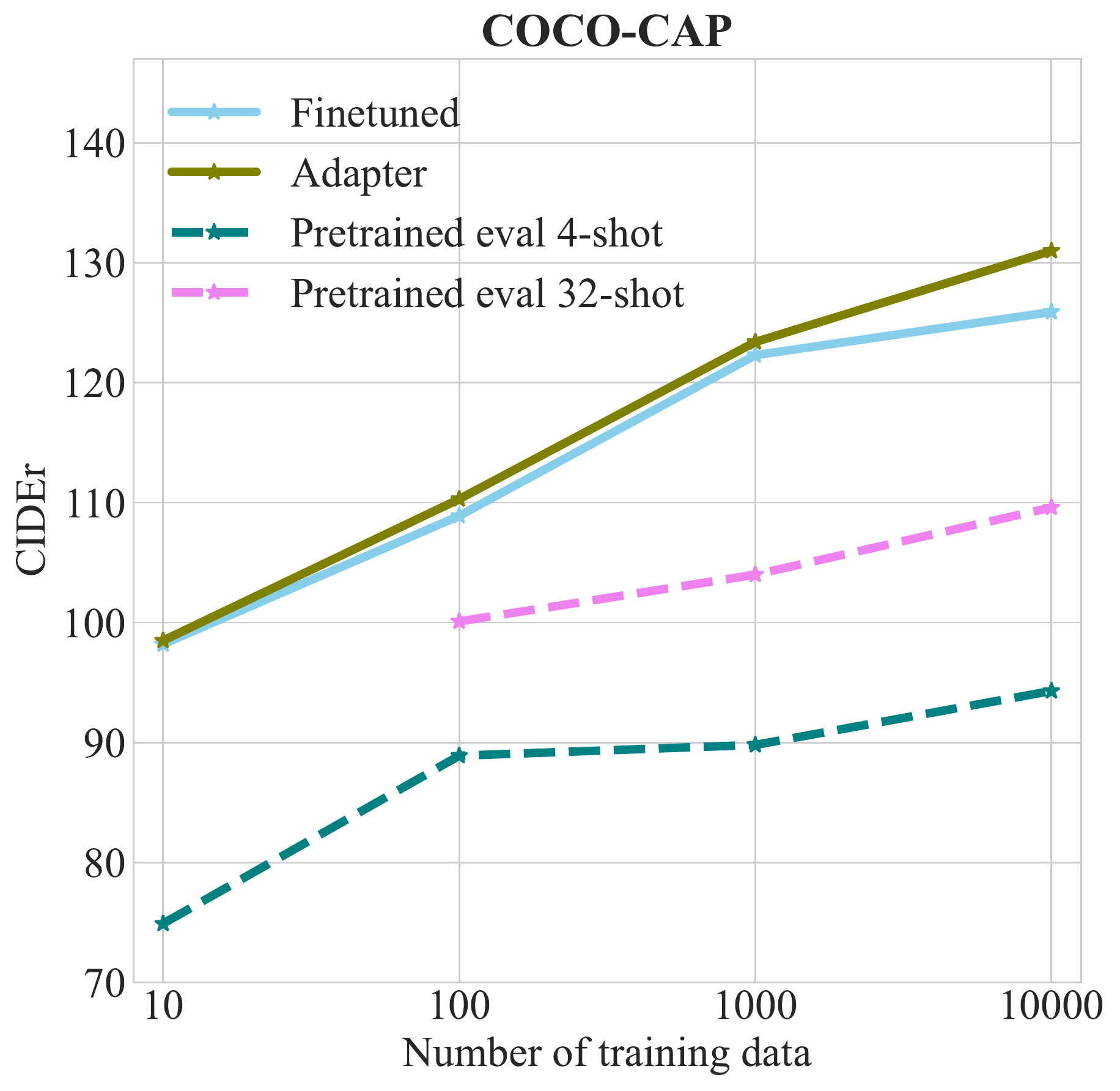}
  \label{fig:coco_ft}
\endminipage\hfill
\minipage{0.25\textwidth}%
  \includegraphics[width=\linewidth]{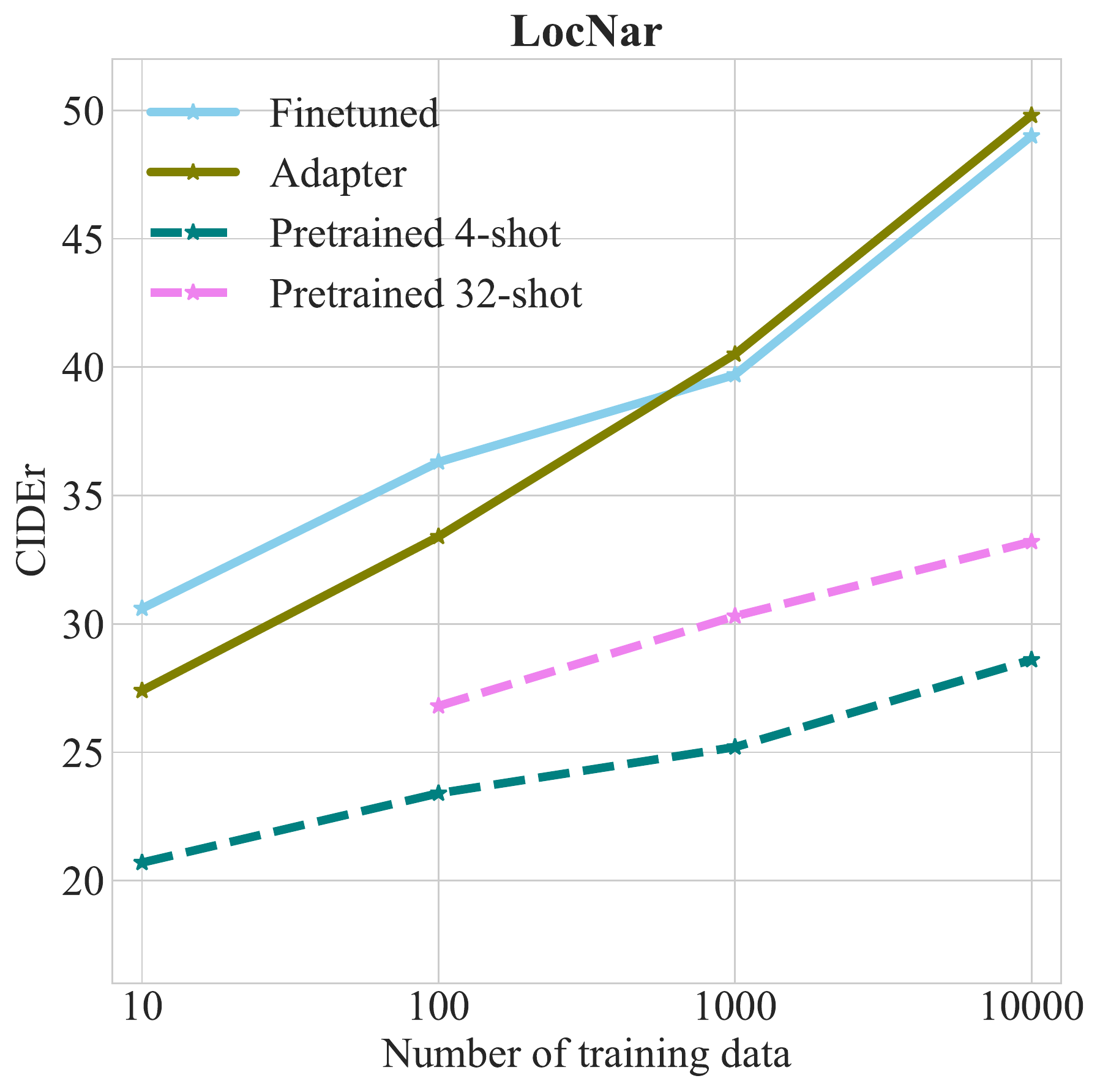}
  \label{fig:LocNar_ft}
\endminipage\hfill
\minipage{0.25\textwidth}
  \includegraphics[width=\linewidth]{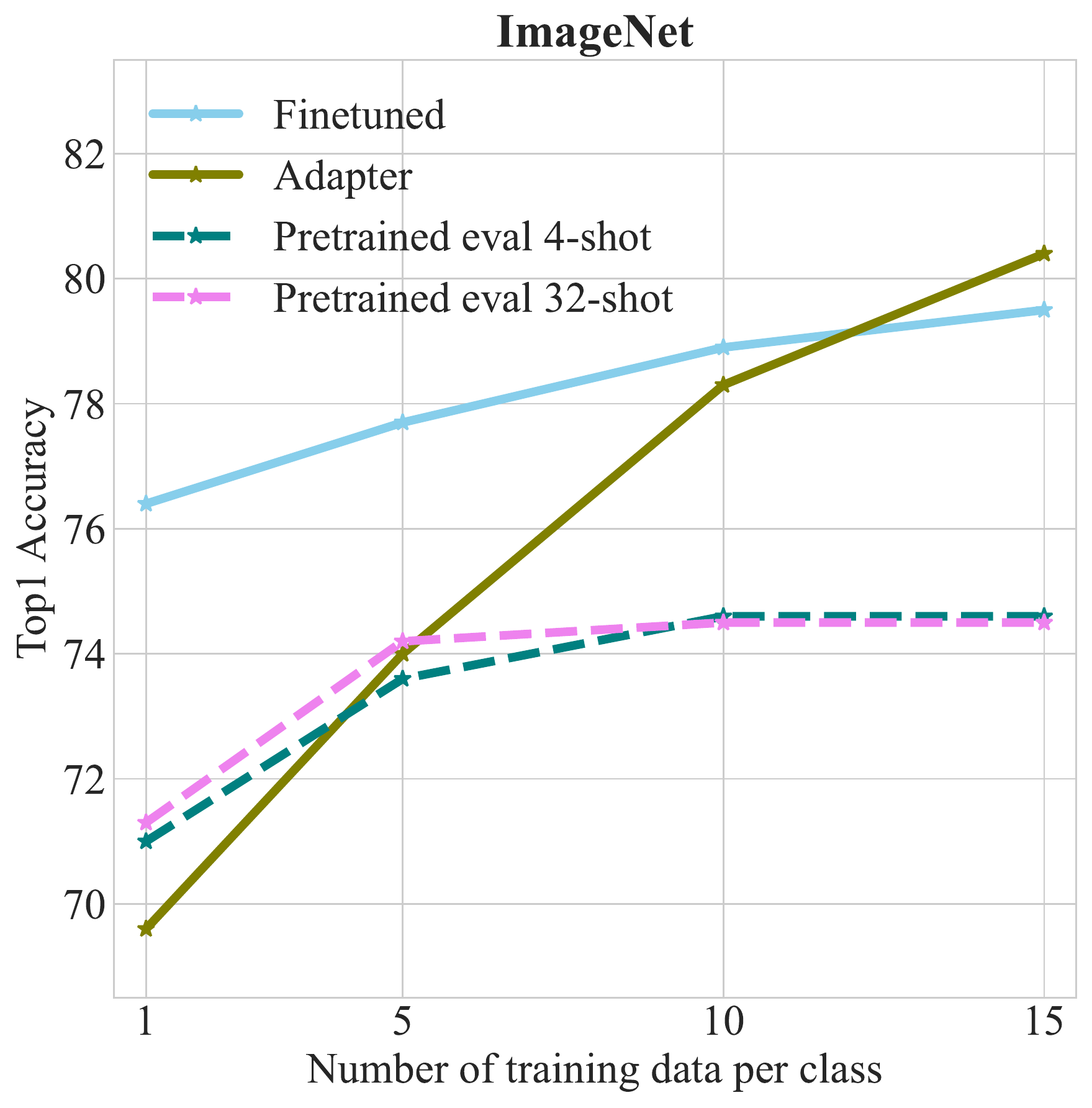}
  \label{fig:IN_ft}
\endminipage\hfill
\minipage{0.25\textwidth}%
  \includegraphics[width=\linewidth]{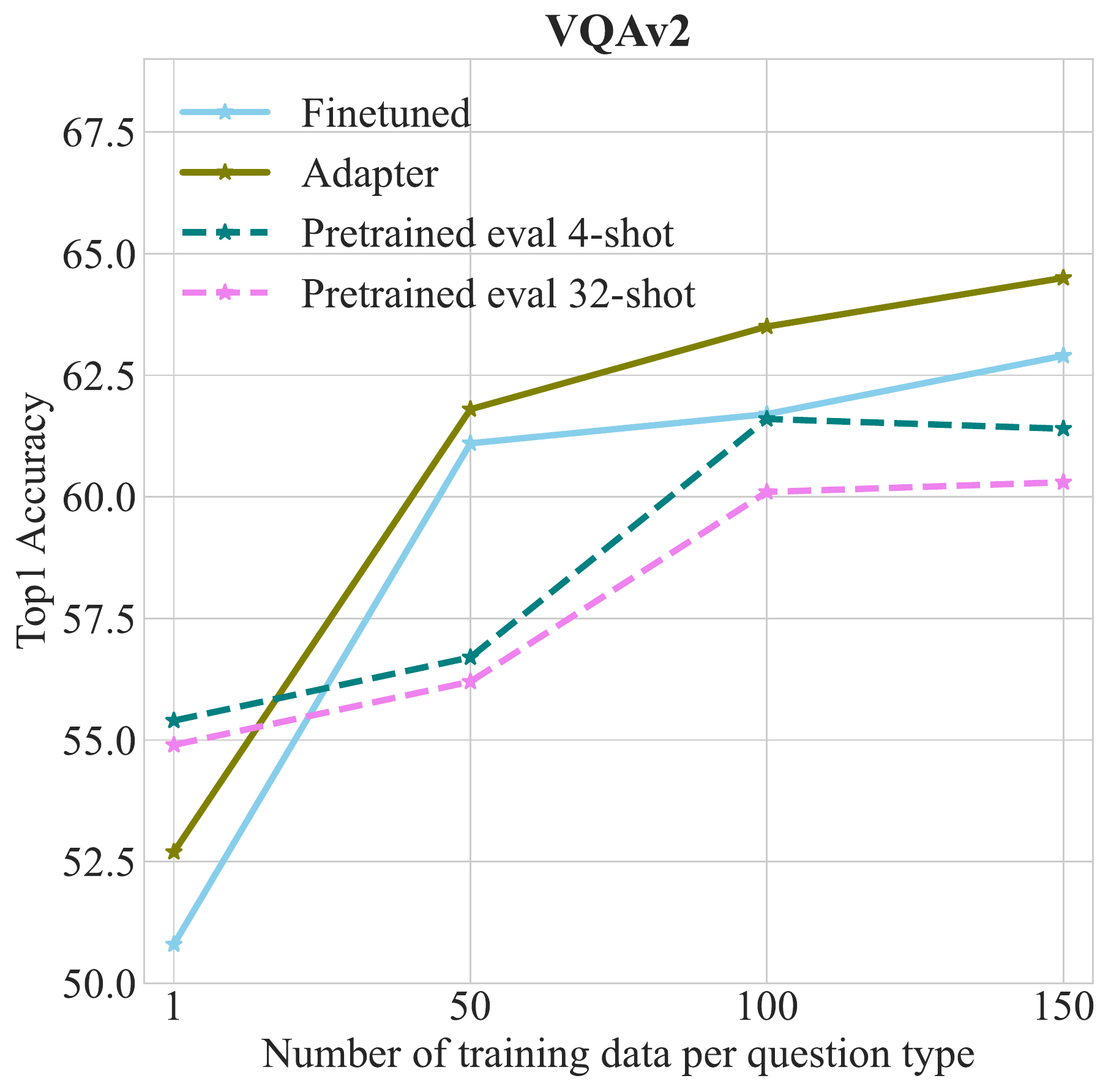}
  \label{fig:vqav2_ft}
 \endminipage
 \vspace{-5mm}
\caption{\textbf{Comparing different ways of training a pseudo-labeller.}  We compare fine-tuning, task adapter and in-context learning as ways to adapt a model in low-data regime on multiple tasks, including captioning (COCO Captioning, Localized Narratives), classification (ImageNet) and VQA (VQAv2). Surprisingly, fine-tuning shows competitive performance on captioning and classification even at the lowest data regime, while in-context learning performs better on VQA when there is only one training sample per question type available. }
 \label{fig:ft}

\end{figure*}

\subsection{Datasets} \label{sec:dataset}
\paragraph{COCO.}
COCO-caption dataset~\cite{cococap2015} contains 330,000 images in total, each one with five independent human generated captions. From the training set, we randomly select 10, 100, 1000, 10000 samples for the low-data regime training, 200 samples for validation. We evaluate on the original validation set.

\paragraph{Localized Narratives.}
LocNar~\cite{locnar2020} is a dataset for conditioned image captioning. It has images with fine-grained text descriptions guided by mouse traces. Different from traditional image captioning with single-sentence captions, its captions are of 36.5 words on average, making it a very novel task to adapt to. We use the Open-Images split in LocNar for our experiments, where we omit the mouse traces and treat it simply as a captioning task.  We randomly select 10, 100, 1000, 10000 samples from the training set for the low-data regime experiments, 200 samples for validation. We evaluate the model on the original test set with 126k samples.

\paragraph{ImageNet.}
ImageNet~\cite{deng2009imagenet} contains 1.2 million annotated images from 1000 classes. We randomly sample 1, 5, 10, 15 images per class for training, and 200 images from each class for validation. We evaluate the model on the original ImageNet validation set with 50k samples.

\paragraph{VQAv2.}
VQAv2~\cite{balanced_vqa_v2} is a dataset containing open-ended questions on 265k images, with 5.4 question per image on average. There are 65 question types, determined by the first few words of the question. From the training set, we randomly sample 1, 50, 100, 150 QA pairs per question type for training, 1024 QA pairs regardless of the question type for validation. We evaluate the model on the original validation with 214K QA pairs.

\subsection{Implementation Details} \label{sec:imple}
\paragraph{Model.}
We use Flamingo-3B~\cite{alayrac2022flamingo} as the main visual language model in our experiments.
For the contrastive model, we use the vision-text contrastive model used in the Flamingo-3B model.
More precisely, to obtain a score between an image and a text candidate, we compute the corresponding vision and text embeddings and use the cosine similarity between the two embeddings as our similarity score.
We discuss this in more details in the filtering experiments in~\cref{sec:ablation}.
Note that the Flamingo-3B model is trained on large-scale image and text pairs (ALIGN~\cite{li2021align} and LTIP~\cite{alayrac2022flamingo}), video and text pairs (VTP) and an interleaved image-text dataset scrapped from webpages (M3W).

\paragraph{Training.}
We train the model for 12 epochs on all the datasets using the AdamW optimizer~\cite{adamw} with global norm
clipping of 1, and decay the learning rate by 0.1 every 4 epochs. The hyper-parameters are chosen on the validation set when the model is trained with 10 samples on COCO.
Total batch size is set to 64 for all the experiments with more than 64 training samples, otherwise, we use batch size 8. The 64 samples are made of equal number of samples from ground-truth data and self-annotated data. Learning rate is set to 7e-6 for fine-tuning, and 1e-4/1e-5 to train the adapter layers from scratch. All hyper-parameters are chosen based on the validation results. All the images are resized to 320x320 with padding, only color jittering is applied as augmentation.

\paragraph{Balance of classes.}
We balance the number of images from each class/question type in both training and validation to ensure all the classes are covered in the few training annotations we sampled. More specifically, we first sample $O$ samples from each class/question type, and randomly sample $N$ from them to form the small training and validation set, where $O \gg N$.  For test we use the original unbalanced test split.

\subsection{Empirical Study} \label{sec:ablation}
\subsubsection{Which technique is the most data-efficient for training our pseudo-labeller?} \label{sec:pseudo-labeller} 

Here we compare different data-efficient techniques for training our pseudo-labeller on four different tasks. 
Specifically, we compare three methods: fine-tuning, learning task adapters and in-context learning. For each method, we carefully choose hyper-parameters (e.g., learning rate, total training steps) on a validation set and report the performance on the original test set in \cref{fig:ft}.

Our most striking finding is that a standard fine-tuning with an early stop performs better or comparably with task adapters and in-context learning in many cases, \emph{even with as few as ten annotated images.}
This is surprising as task adapters and in-context learning have been designed to work better in this low-data regime. 

The only exception is VQAv2, where in-context learning performs the best given 1 training sample per question type.
In this setting, fine-tuning may be biasing the model's output distribution towards the answer that was randomly drawn for a given question type. 
Fine-tuning is furthermore known to exploit spurious correlations and biases present in the training data; an issue which may be mitigated by in-context learning~\cite{brown2020language}. In particular, fine-tuning may hence be more sensitive than in-context learning to the mismatch between the uniform distribution of question type during training versus at test time.

Overall the performance of task adapter and fine-tuning is better than in-context learning even when only a few samples are available.

To carry out the self-labelling experiments, we use the best training technique in the smallest data-regime. In particular, we use a 4-shot in-context learning adapted model as the pseudo-labeller on VQAv2, and a fine-tuned model as the pseudo-labeller on COCO, ImageNet and LocNar.

\subsubsection{What is the best way of training with pseudo-labels?}
In the following experiments, we focus on semi-supervised training with pseudo-labels and GT labels. We conduct the experiments on three different tasks, one dataset each -- COCO for captioning, ImageNet for classification and  VQAv2 for VQA. We omit Localized Narratives for ablations because it falls into the same task category as COCO.

\paragraph{On the importance of the weighting of the pseudo-labelled data}
\begin{table}[t]
\centering
\resizebox{0.4\textwidth}{!}{%
\begin{tabular}{cccc}
\hline
\textbf{\begin{tabular}[c]{@{}c@{}}Groundtruth \\ weight $\alpha$ \end{tabular}} &
  \textbf{\begin{tabular}[c]{@{}c@{}}ImageNet \\ Top1\end{tabular}} &
  \textbf{\begin{tabular}[c]{@{}c@{}}COCO \\ CIDEr\end{tabular}} &
  \textbf{\begin{tabular}[c]{@{}c@{}}VQAv2 \\ Top1\end{tabular}} \\ \hline
0  &  $70.7_{\pm{0.9}}$      & $105.1_{\pm0.3}$          & $53.7_{\pm1.8}$   \\\hline
0.1 & $\textbf{76.2}_{\pm{0.9}}$ & $\textbf{105.4}_{\pm2.4}$  & $53.0_{\pm1.1}$  \\
0.3 & $74.8_{\pm0.6}$          & $103.1_{\pm2.6}$        &  $52.7_{\pm2.2}$ \\
0.5 & $75.4_{\pm1.1}$         & $103.3_{\pm1.7 }$         &  $\textbf{54.2}_{\pm2.5}$ \\
0.7 & $75.2_{\pm1.0}$          & $101.9_{\pm0.7}$          &  $53.9_{\pm2.7}$\\
0.9 & $74.8_{\pm1.1}$          & $101.8_{\pm0.8 }$        &  $51.3_{\pm1.3}$  \\ \hline
1.0 & $75.9_{\pm1.4}$        & $98.2_{\pm1.2}$          &  $52.3_{\pm0.6}$ \\ \hline
\end{tabular}%
}
\caption{\textbf{On the importance of correctly weighting the ground-truth data against the pseudo generated ones.}}
\label{tab:task_weight}
\end{table}

 We label 10k pseudo-labels on COCO, ImageNet and VQAv2, and
ablate the weight $\alpha$ (\cref{eq:cotrain_loss}) which measures the importance of the manually annotated examples against the pseudo-generated ones.
We conduct our experiments in the lowest data-regime (10 images in captioning, 1 image per class in classification, 1 image per question type in VQA) for first-stage adaptation. We choose the top 10\% pseudo-labels with the highest likelihood for the semi-supervised fine-tuning.
We run the training five times with different random seeds and report the averaged results to reduce the variance. 
The results are shown in \cref{tab:task_weight}, where two main findings emerge.
First, it is not sufficient to only-train on the pseudo-generated labels ($\alpha=0$), ground-truth labels still serve as a source of regularization for good performance.
Second, a careful weighting ($\alpha=0.1$ for ImageNet and COCO, $\alpha=0.5$ for VQAv2 ) is crucial to obtain the best performance on all the datasets. 

\begin{figure*}[h!]
\minipage{0.34\textwidth}
  \includegraphics[width=\linewidth]{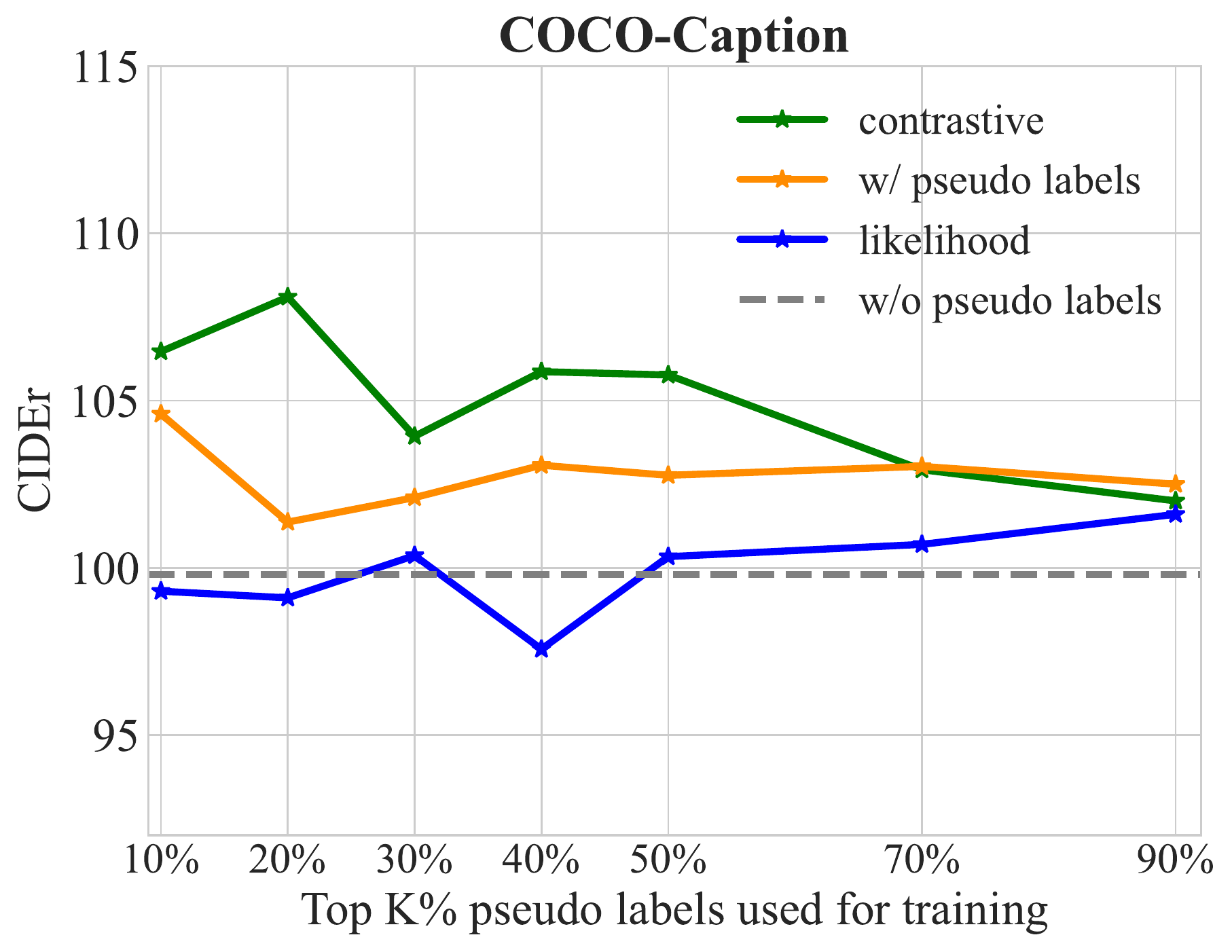}
  \label{fig:coco_filter}
\endminipage
\minipage{0.33\textwidth}
  \includegraphics[width=\linewidth]{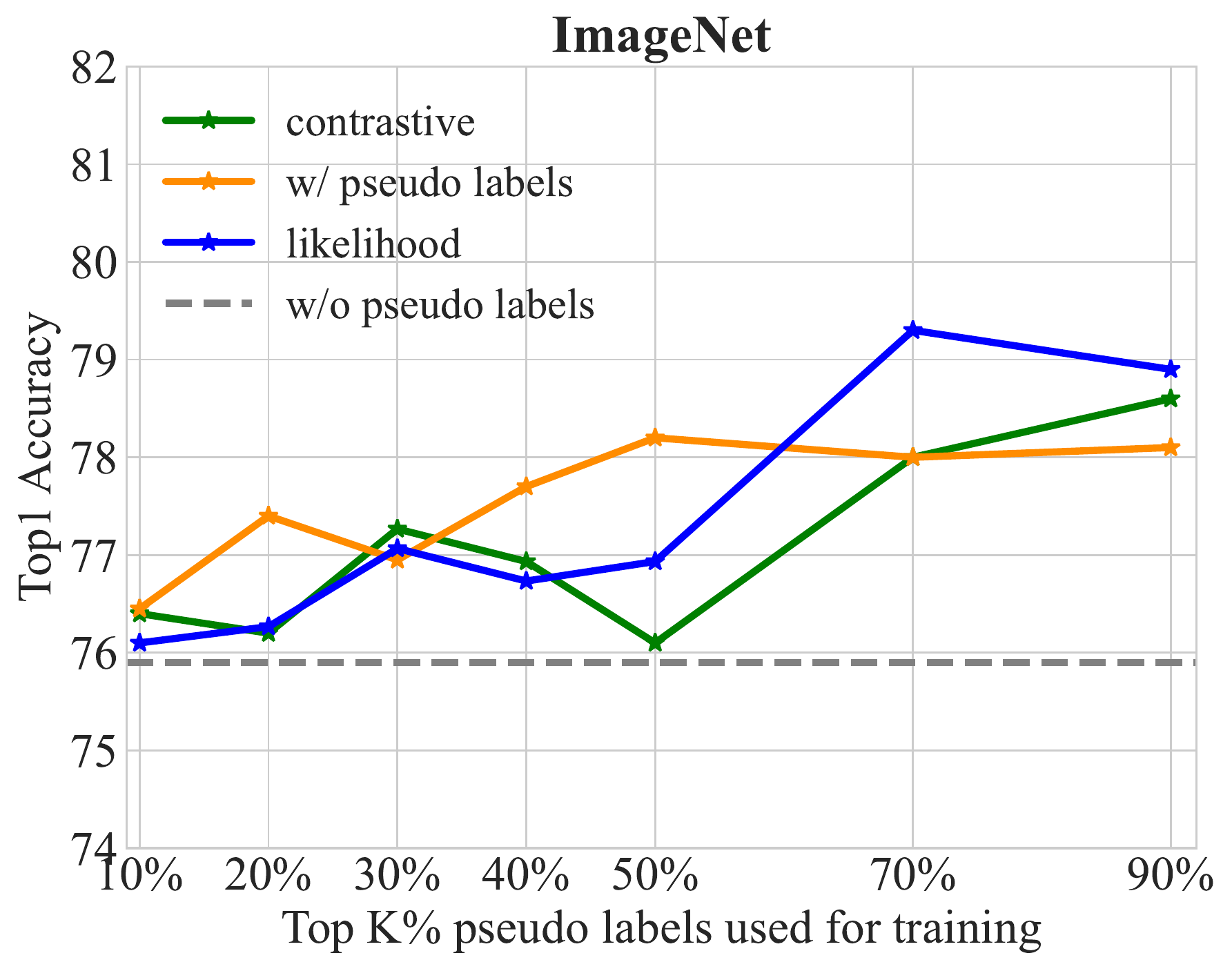}
  \label{fig:IN_filter}
\endminipage\hfill
\minipage{0.33\textwidth}
  \includegraphics[width=\linewidth]{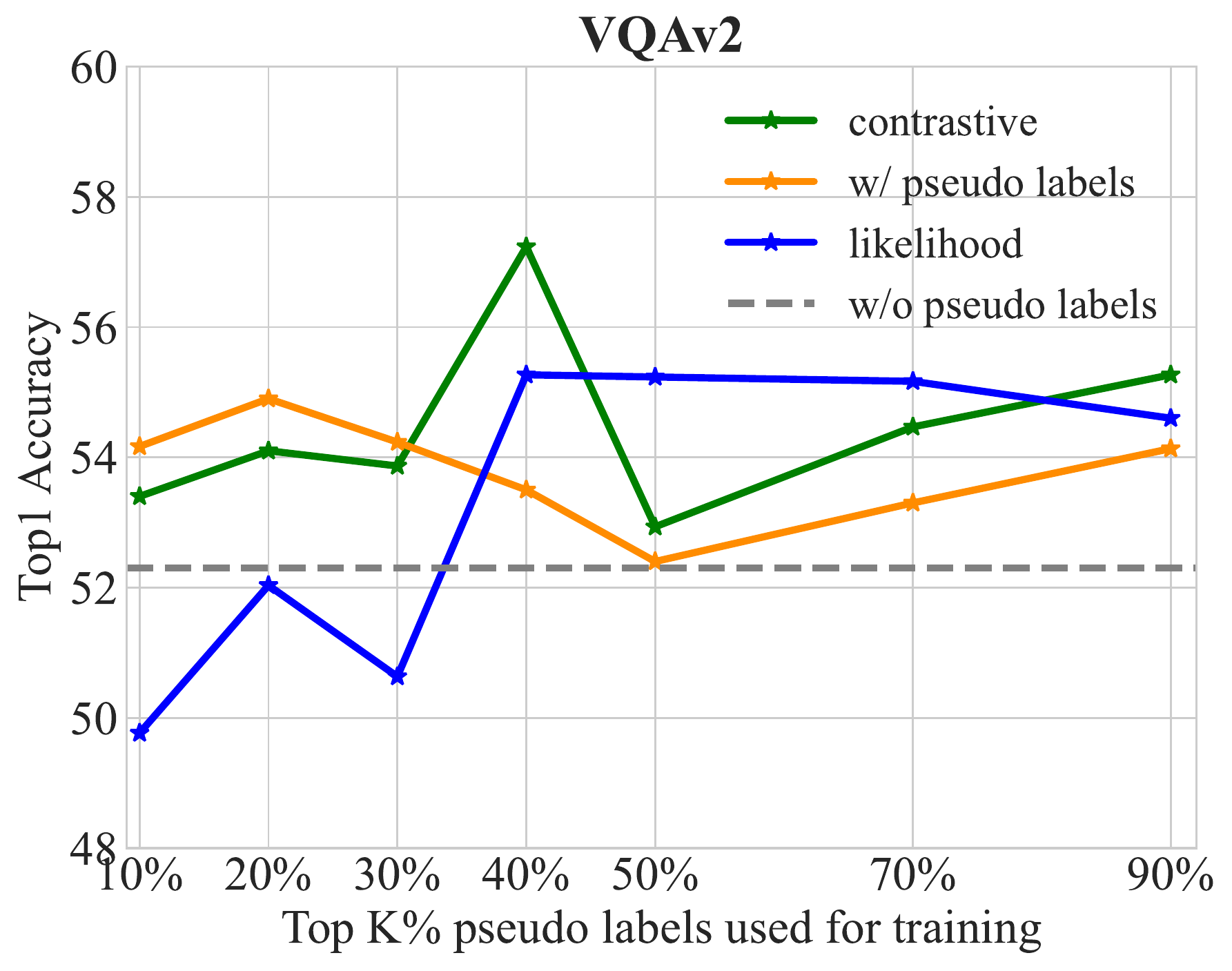}
  \label{fig:IN_filter}
\endminipage\hfill
\vspace{-5mm}
\caption{\textbf{Comparing different filtering methods on captioning, classification, VQA.} We randomly generate 10k pseudo labels for each dataset, and conduct semi-supervised learning with top p\% pseudo-labels filtered based on contrastive scores and likelihood respectively on the three tasks. }
\label{fig:filtering}

\end{figure*}

\paragraph{Does more pseudo-labels help? }
We show the performance of training with an increasing number of pseudo-labels in \cref{fig:filtering}. We labelled 10k images on each dataset, keep the weight ratio fixed and randomly sample a certain percentage of them for semi-supervised training, ranging from 10\% to 90\% (1k to 9k pseudo-labels). Results show that having more unfiltered labels does not have a huge impact on open-ended tasks like COCO and VQAv2, but it brings performance gains on close-ended classification on ImageNet. 

\paragraph{Does filtering help?}

On the three tasks, we explore the following filtering strategy: 
\begin{itemize}
\item \textbf{Contrastive filtering.} Given an image and the top 3 predictions from beam search (beam width=3), we use a contrastive model to compute the similarity scores between each image and the top 3 captions. Based on the score, we first choose the best matched caption for each image to form an image-text pair, and then select the top K\% image-text pairs for training. 

\item \textbf{Likelihood filtering.} Given an image and the top 3 predictions from beam search (beam width=3), we choose the caption with maximum likelihood to form an image-text pair, and then select the top K\% image-text pairs for training.\\
\end{itemize}
In \cref{fig:filtering}, we show the results of the above filtering methods compared to randomly sampling results without filtering. As a baseline, the performance of the pseudo-labeller which has only been trained on ground-truth labels is also shown as dashed grey line in the same plot. On all the three tasks, training with pseudo-labels helps to improve the performance even without filtering. On COCO, contrastive filtering improves the captioning performance. The downward curves shows that the more strict the filtering is, the better the results are. Likelihood filtering worsens the results on captioning, likely due to the fact that likelihood favors short captions. However, contrastive filtering does not bring any significant gain on the other tasks. This might due to the fact that the predictions from it are highly correlated with the predictions from Flamingo, since the visual backbone of Flamingo and the contrastive model are pre-trained on the same data.

For likelihood filtering, although overall it does not lead to the best performance, the performance goes up when we increase the number of samples from top 10\% to 70\% on all the datasets, which is probably the results of two components: 1) training on samples with relatively low confidence helps the model to learn new things,  2) labels generated by large pre-trained VLM are sufficiently clean, so that there are enough samples with low confidence but containing correct and useful information.

\paragraph{Does more pseudo-labels of better quality help? }
In order to see whether an increasing number of pseudo-labels of good quality helps, we use the pseudo-labeller to label more images and set a certain threshold for selection.   The experiment is conducted on COCO captioning as this is the dataset where filtering shows a consistent and obvious improvement. We set the threshold of contrastive scores as 0.2, and use 100, 1000 and 10000 of them for semi-supervised learning. Results in \cref{tab:filtered_coco} shows that increasing the number of labels from 100 to 1000 helps improve the CIDEr score by 2.6, but the performance saturates when we increase the number from 1000 to 10000.

\begin{table}[h]
\centering
\renewcommand{\arraystretch}{1.2}
\resizebox{0.85\linewidth}{!}{%
\begin{tabular}{cc}\hline
\begin{tabular}[c]
{@{}c@{}}\textbf{Number of pseudo-labels above threshold}\\ (contrastive score \textgreater 0.2)\end{tabular} & \textbf{CIDEr} \\ \hline
\textbf{100}   & $103.9_{\pm1.2}$      \\
\textbf{1000}  & $106.5_{\pm0.7}$ \\
\textbf{10000}& $\textbf{106.6}_{\pm0.3}$      \\ \hline
\end{tabular}%
}
\caption{\textbf{Semi-supervised learning with an increasing number of high-quality filtered data on COCO captioning.} We use 100, 1000, 10000 pseudo-labels with contrastive scores higher than 0.2 and investigate whether more clean pseudo-labels further help the performance. }
\label{tab:filtered_coco}
\end{table}

\begin{figure}[b]
\includegraphics[width=\linewidth]{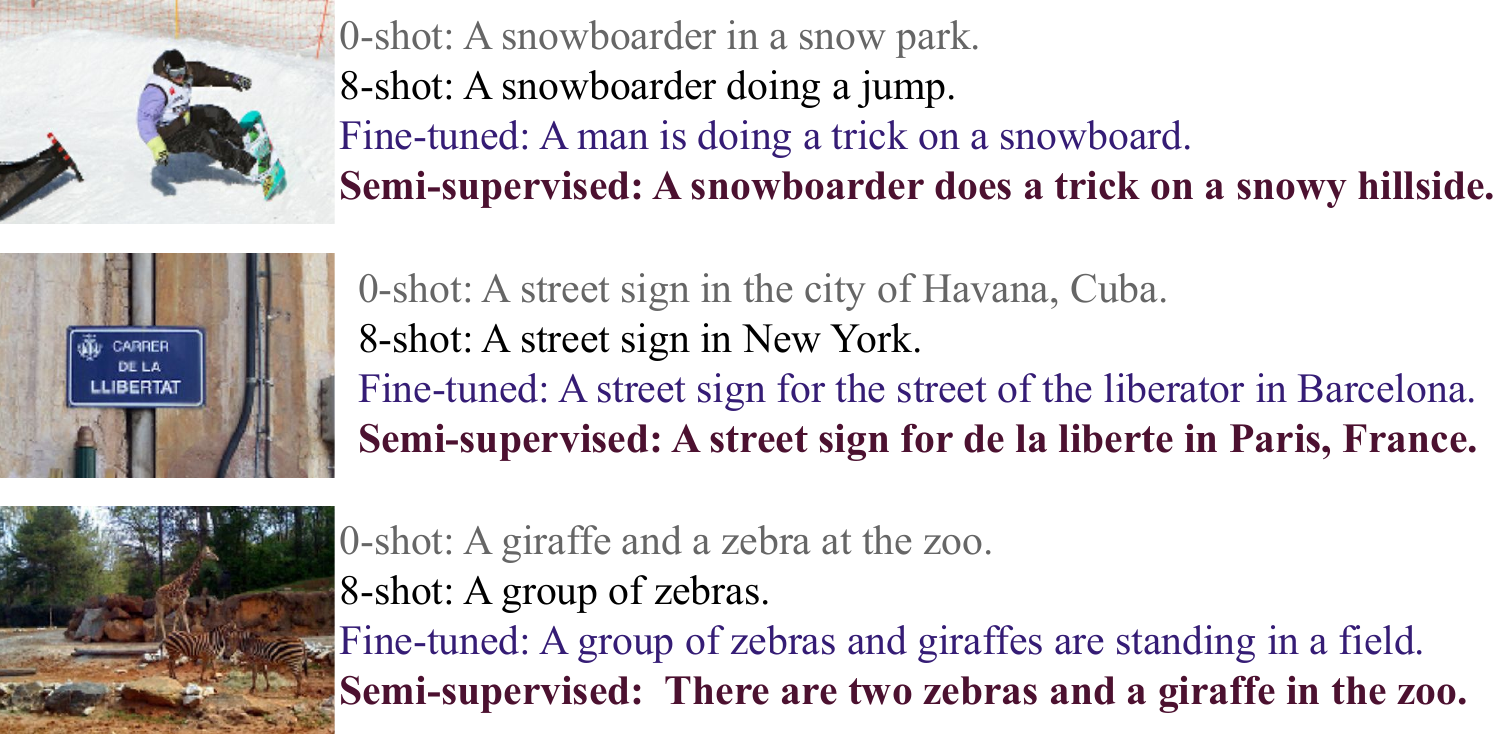}
\vspace{-5mm}
\caption{\textbf{Visualizations of predictions from different models on COCO.} Fine-tuning and self-labelling improve the quality of the captions predicted, and generate more detailed and precise captions. }
\end{figure}

\subsection{How much can we learn from 10 samples?} \label{sec:results}
On all the four datasets, we take the best model of semi-supervised based on the results on validation set, and evaluate them on the test set. As shown in \cref{tab:results}, semi-supervised learning brings significant improvement on all the tasks. The Flamingo 3B model with self-labelling achieves 110.7 CIDEr score on COCO captioning, and Top1 accuracy 78.3 on ImageNet classification. This is about the same or better than a Flamingo 80B model prompted by 16 samples (110.5 on COCO and 71.7 on ImageNet). Our results show that even a few samples can adapt the model to a new task efficiently and effectively.
\begin{table}[]
\renewcommand{\arraystretch}{1.5}
\centering
\resizebox{\linewidth}{!}{%
\begin{tabular}{cccccc}
\hline
\textbf{Model}                         & \textbf{Method}        & \textbf{COCO}  & \textbf{ImageNet} & \textbf{VQAv2} & \textbf{LocNar} \\ \hline
                                       & \textbf{Fine-tuning}   & 98.2          & 76.4              & 50.8           & 30.6                \\
                                       & \textbf{Adapter}       & 98.5           & 69.6              & 52.7           & 27.4            \\
                                       & \textbf{8-shot}        & 77.9           & 71.5              &  55.4         & 27.3            \\

\multirow{-5}{*}{\textbf{Flamingo 3B}} & \textbf{Self-labelling} & \textbf{110.7} & \textbf{78.3}     & \textbf{58.0}      & \textbf{31.0}       \\ \hline
{\color[HTML]{656565} \textbf{Flamingo 9B}} &
  {\color[HTML]{656565} \textbf{8-shot}} &
  {\color[HTML]{656565} 99.0} &
  {\color[HTML]{656565} 71.2} &
  {\color[HTML]{656565} 58.0} &
  {\color[HTML]{656565} 31.7} \\
  & 
    {\color[HTML]{656565} \textbf{16-shot}} & 
  {\color[HTML]{656565}102.2} & 
  {\color[HTML]{656565}59.4} & 
  {\color[HTML]{656565}71.7 } &
  {\color[HTML]{656565} 34.5} \\ \hline
{\color[HTML]{656565} \textbf{Flamingo 80B}} &
  {\color[HTML]{656565} \textbf{8-shot}} &
  {\color[HTML]{656565} 108.8} &
  {\color[HTML]{656565} 71.9} &
  {\color[HTML]{656565} 65.6} &
  {\color[HTML]{656565}\textbf{38.4}} \\
{\color[HTML]{656565} \textbf{}} &
  {\color[HTML]{656565} \textbf{16-shot}} &
  {\color[HTML]{656565}110.5} &
  {\color[HTML]{656565} 71.7 } &
  {\color[HTML]{656565}\textbf{66.8}} &
  {\color[HTML]{656565} 37.5 } \\ \hline
\end{tabular}%
}
\vspace{-2mm}
\caption{\textbf{Semi-supervised learning with self-labelling brings improvement on all the four datasets.} We train the models on the fewest training data (10 samples for captioning, 1 sample per class per classification, 1 sample per question type for VQA) and report results on the test set. }
\label{tab:results}
\vspace{-5mm}
\end{table}

\section{Conclusion}

In this work we have explored how to best exploit a few annotated datapoints to adapt a pre-trained visual language model to a wide variety of tasks (object classification, image captioning, visual question answering).
Next, we summarize what we have learned as well as the future directions.

In the setting of our work, we learned three things that we found surprising and deem worth of sharing:
\begin{enumerate}
    \item \textbf{Best Adaptation Method is task-dependent.} When adapting a model to a new task with only a few labels (even just 10 examples), we have found that none of the standard methods - fine-tuning, adapter or in-context learning - works consistently well on all the datasets. Fine-tuning works better in captioning and classification, while in task like VQA, it performs worse than in-context learning as it is more likely learn spurious correlations between training samples and the answers.
 
    \item \textbf{Self-labelling using a effectively adapted self-labeller can bring significant gain across all tasks.} Across the 3 tasks and 4 datasets considered, using self-labelling always brought significant benefits. However, one may need to adapt which techniques to use for producing the pseudo-labels depending on the nature of the task.
    
    \item \textbf{Self-labelling works even without carefully designed filtering and heavy augmentation when using large pre-trained model.} Interestingly, we found out that self-labelling brings improvement without filtering and heavy augmentation. This is in contrast to previous self-labelling work which can be done with model trained on data of much smaller scale~\cite{sohn2020fixmatch,noisystudent2020,blip2022}. 
    With task-specific, well-designed post-processing method (e.g., filtering), we find that the performance of self-labelling can be further improved.

\end{enumerate}

\paragraph{Limitations and Future Work.}
First, we note that our method still require a validation set (200 images in practice) in order to select the right adaptation methods and hyper-parameters.
In practice, this set is effectively part of the training pool.
Finding ways to use even less validation data, or make sure the hyper-parameters can transfer across tasks is an important endeavor that we leave for future work.

Finally, we have only worked with models up to 1.3B scale, but would be keen to see how these learnings hold at larger scale.
Overall, we believe that our work opens up new avenues to improve visual language models in the presence of few annotated data.

{\small
\bibliographystyle{ieee_fullname}
\bibliography{egbib}
}

\clearpage
\appendix
\noindent{\Large{\textbf{Appendix}}}

\section{Implementation Details}
\subsection{Details on Fine-tuning}
\begin{figure}[h]
\centering
\includegraphics[width=\linewidth]{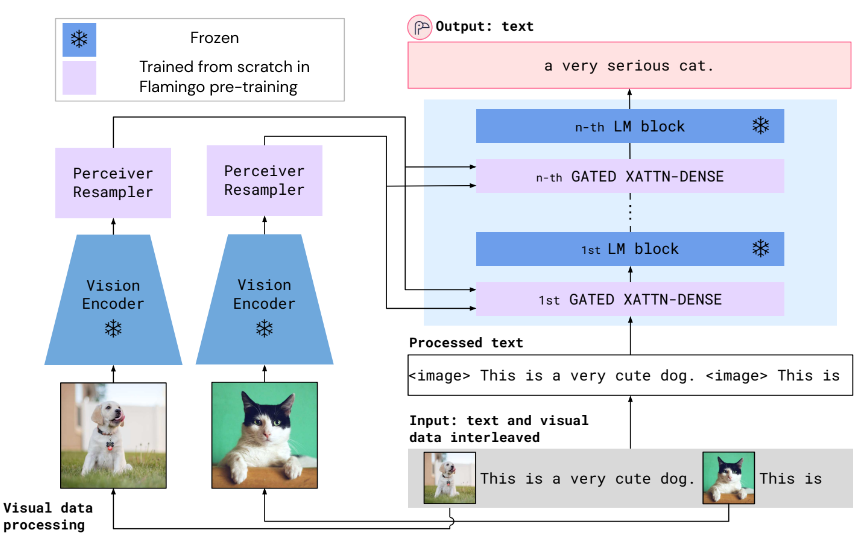}
\caption{\textbf{Frozen and unfrozen modules in Flamingo (figure reproduced from \cite{alayrac2022flamingo})} We only fine-tune the modules which are trained in the original pre-training while keep the others frozen.}
\label{fig:finetune}
\end{figure}

Following the training settings in Flamingo 3B~\cite{alayrac2022flamingo}, we fine-tune the same modules which are trained during pre-training. In more details and as illustrated in \cref{fig:finetune}, we keep the vision encoder and LM blocks frozen, and finetune the Perceiver Resamples and Gated XATTN-DENSE blocks. In total, there are 1.4B parameters being fine-tuned.

\subsection{Details on Task Adapter}

\begin{figure}[h]
\centering
\includegraphics[width=\linewidth]{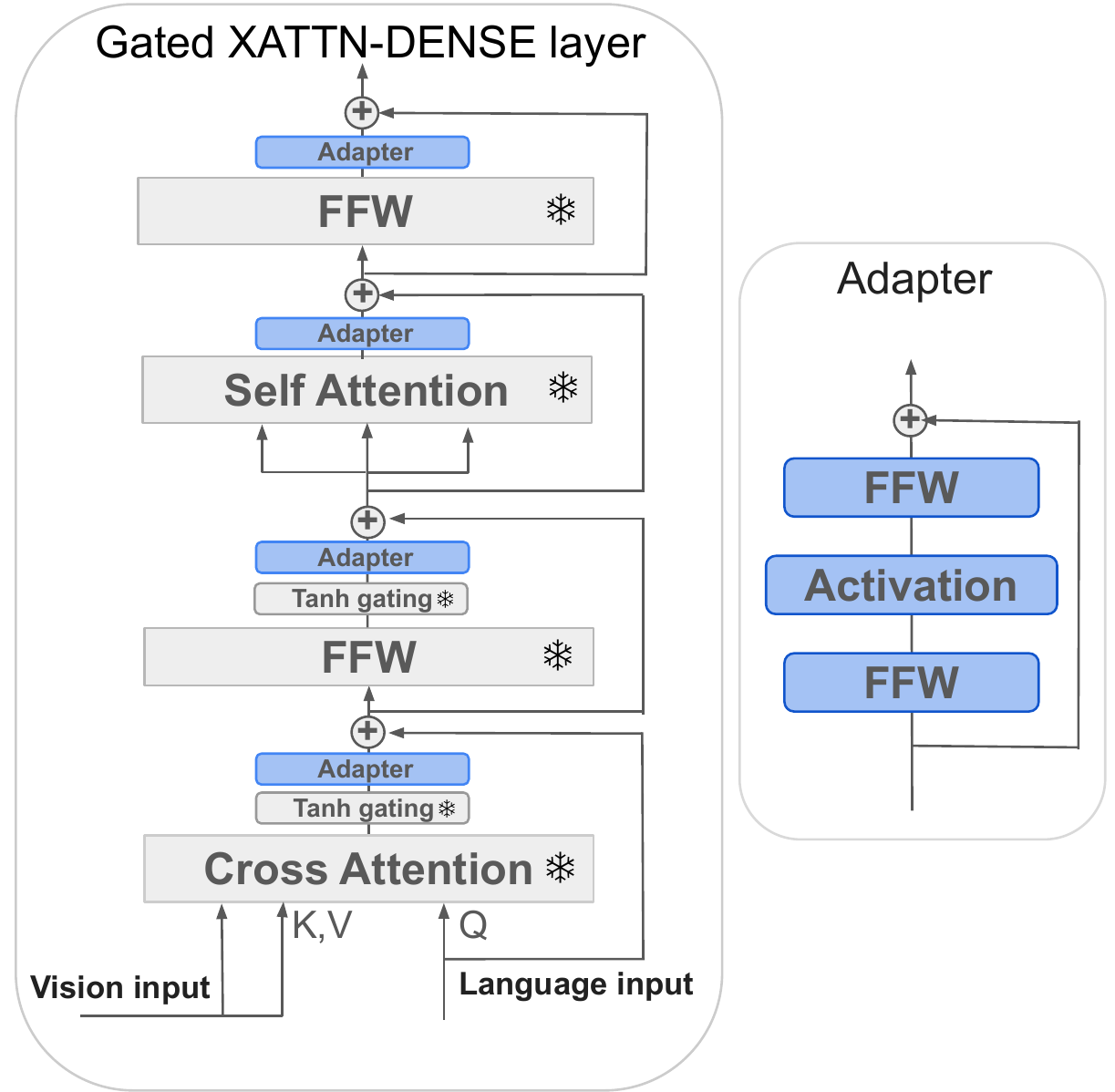}
\caption{\textbf{Inserting Task adapter layers in Flamingo.} \textbf{Left:} we add four adapters in every block of the Flamingo visual-language model decoder. \textbf{Right:} Each adapter is a MLP made up of two feed-forward layers and one activation layer in between.}
\label{fig:adapter}
\end{figure}

We add task adapters~\cite{adapter2019} after every Cross-Attention, Self-Attention and all feed-forward layers in the Language Model stack, as shown in \cref{fig:adapter}. 
We fine-tune the adapters and the layer normalization parameters~\cite{ba2016layer} in the model, resulting in 13M parameters being fine-tuned.

\section{Self-labelling on Out-of-distribution images}
In the main paper, we conduct all the self-labelling experiments on in-distribution image based on the assumption that it is easy to get unlabelled images from the same distribution in real life. For examples, when a zoologist wants to adapt the VLM to images of wild life, it is easy to collect hundreds of unlabelled images and manually label 10 out of them. 
\begin{figure*}[h!]
\centering
\includegraphics[width=0.95\linewidth]{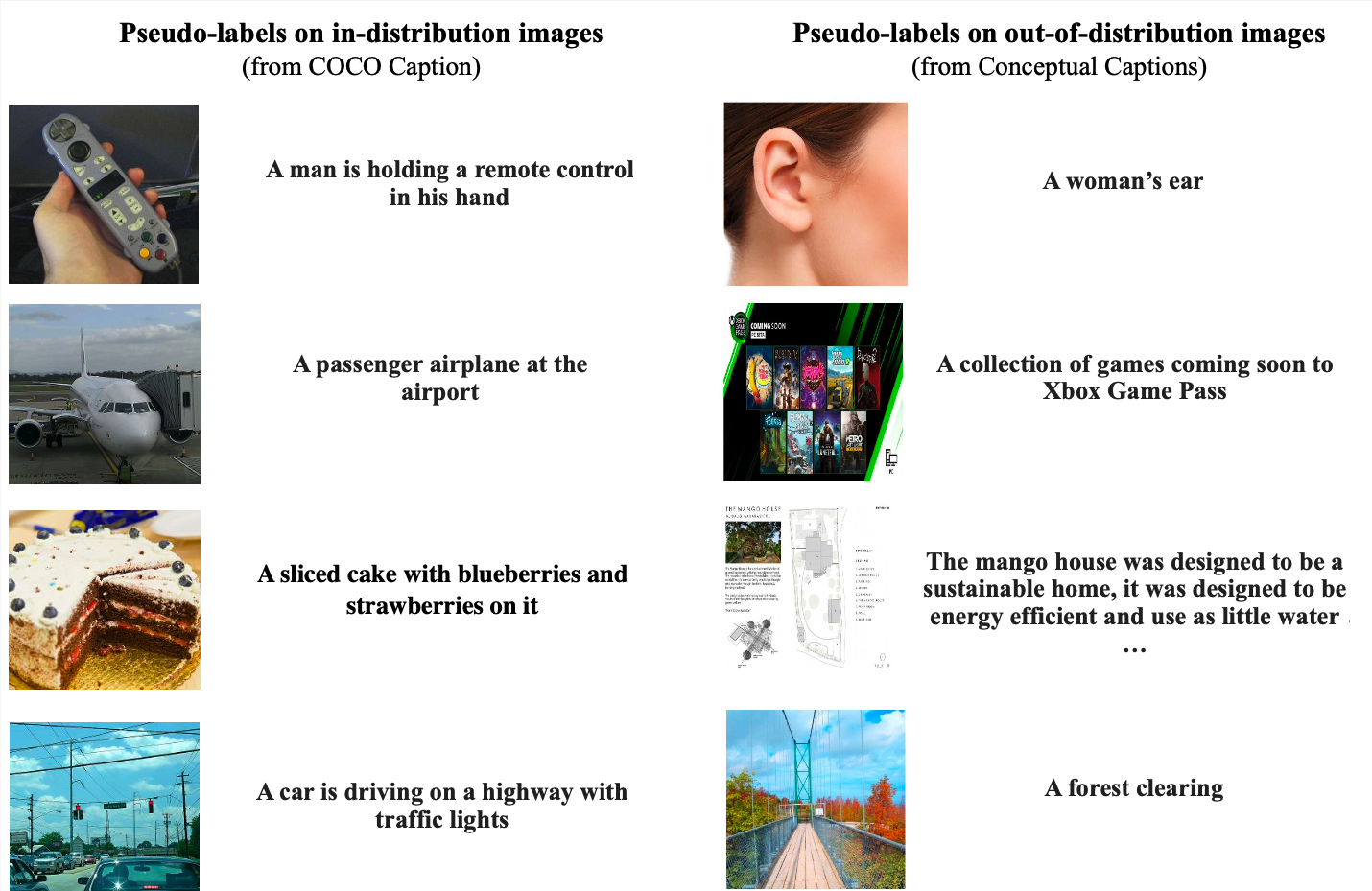}
\caption{\textbf{Pseudo-labels generated on in-distribution images and out-of-distribution images.} Using a Flamingo 3B trained on 10 samples from COCO caption, on images from both COCO-Caption(left) and Conceptual Captions (right). }
\label{fig:ood}
\end{figure*}

To see the results by using out-of-distribution (OOD) images, we repeat the self-labelling pipeline but on unlabeled images from Conceptual Captions 12M~\cite{changpinyo2021cc12m}. First, the VLM model is fine-tuned on 10 image-text pairs from COCO Caption. It is then used to generate pseudo labels on CC12M, which are added to the 10 GT samples from COCO as the training data in semi-supervised learning stage. Training with OOD images leads to 100.6  CIDEr score on the final test split, which is lower than plain fine-tuning by 0.5. It might be due to the fact that given images that look different from the 10 samples in fine-tuning, the model tends to generate captions which are more similar to the ones in pre-training, some examples of pseudo captions generated in both cases are shown in \cref{fig:ood}. 

\section{Qualitative Results on VQAv2}
\begin{figure*}[t]
\centering
\includegraphics[width=\linewidth]{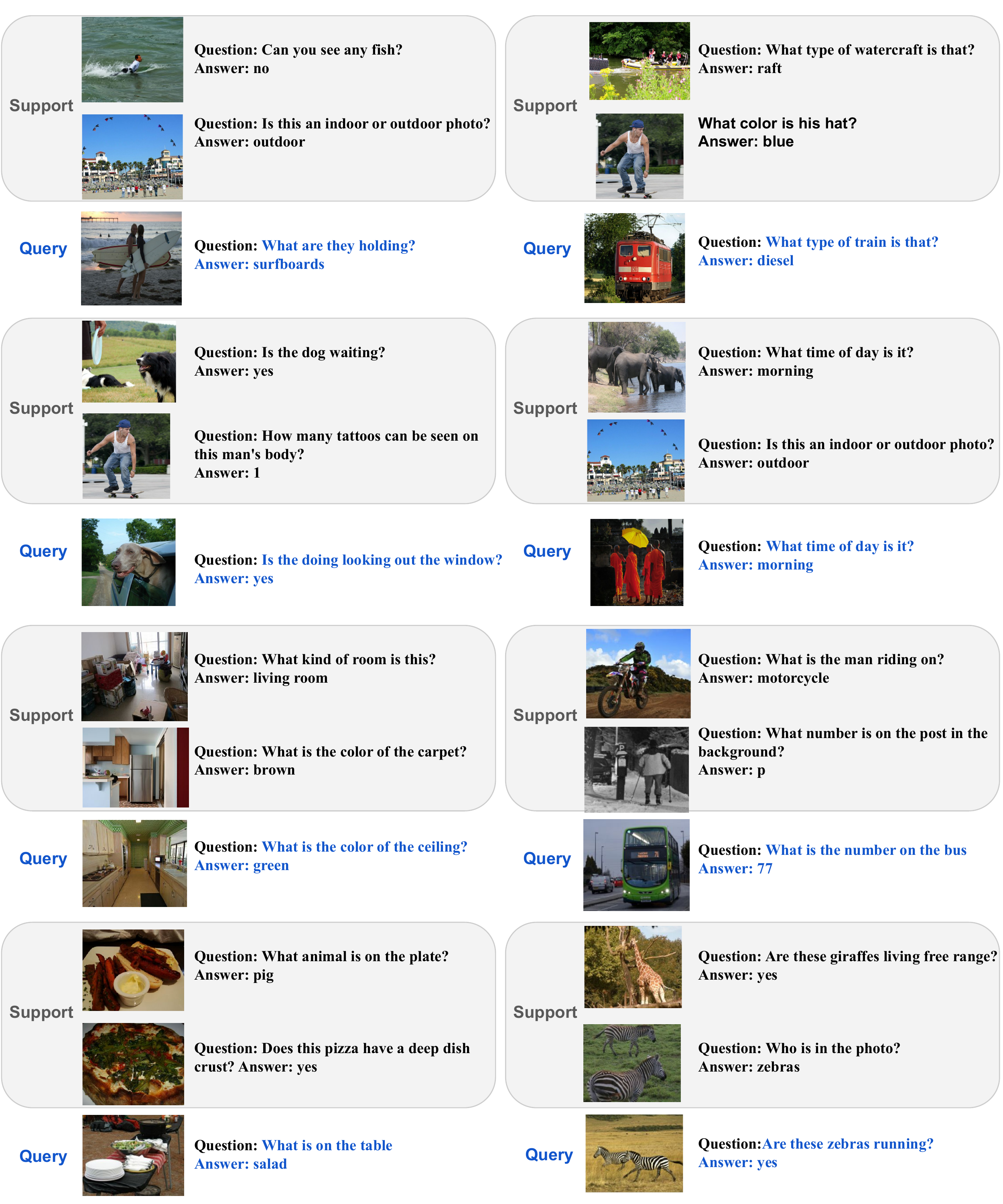}
\caption{\textbf{Generated Question and Answers through in-context learning in VQAv2.} We 4-shot prompt a flamingo 3B model to generate questions and answers on unlabelled images. Support images and texts are shown in black, query images and texts are shown in blue. Only two out of four support images are shown for each query for illustration purpose.}
\label{fig:generated_qa}
\end{figure*}

In \cref{fig:generated_qa} we show more examples of questions and answers generated in self-labelling on VQAv2, with 4-shot prompted Flamingo 3B. Only two out of four support images are shown for each query for illustration purpose.

\end{document}